\documentclass[twoside,11pt]{article}

\usepackage{jmlr2e}
\usepackage{epsfig}
\usepackage{graphicx}
\usepackage{amsmath}
\usepackage{amssymb}
\usepackage{lipsum,lineno}
\usepackage{textcomp}
\usepackage{hyperref}

\jmlrheading{x}{2015}{x-xx}{5/15}{xx/xx}{Hoo-Chang Shin, Le Lu, Lauren Kim, Ari Seff, Jianhua Yao, Ronald M. Summers}

\ShortHeadings{Text/Image Mining on a Large-Scale Radiology Database}{Hoo-Chang Shin, Le Lu, Lauren Kim, Ari Seff, Jianhua Yao, Ronald M. Summers}
\firstpageno{1}

\begin{document}

\title{Interleaved Text/Image Deep Mining on a Large-Scale Radiology Database for Automated Image Interpretation}

\author{\name Hoo-Chang Shin \email hoochang.shin@nih.gov \\
        \name Le Lu \email le.lu@nih.gov \\
        \name Lauren Kim \email lauren.kim2@nih.gov \\
        \name Ari Seff \email ari.seff@nih.gov \\
        \name Jianhua Yao \email jyao@cc.nih.gov \\
        \name Ronald M. Summers \email rms@nih.gov \\
        \addr Imaging Biomarkers and Computer-Aided Diagnosis Laboratory\\
        Radiology and Imaging Sciences\\
        National Institutes of Health Clinical Center\\
        Bethesda, MD 20892-1182, USA
       }

\editor{xxx xxxx}

\maketitle

\begin{abstract}
Despite tremendous progress in computer vision, there has not been an attempt for machine learning on very large-scale medical image databases.
We present an interleaved text/image deep learning system to extract and mine the semantic interactions of radiology images and reports from a national research hospital's Picture Archiving and Communication System. 
With natural language processing, we mine a collection of representative $\sim$216K two-dimensional key images selected by clinicians for diagnostic reference, and match the images with their descriptions in an automated manner.
Our system interleaves between unsupervised learning and supervised learning on document- and sentence-level text collections, to generate semantic labels and to predict them given an image.
Given an image of a patient scan, semantic topics in radiology levels are predicted, and associated key-words are generated.
Also, a number of frequent disease types are detected as present or absent, to provide more specific interpretation of a patient scan.
This shows the potential of large-scale learning and prediction in electronic patient records available in most modern clinical institutions.
\end{abstract}

\begin{keywords}
  Deep learning, Convolutional Neural Networks, Topic Models, Natural Language Processing, Medical Imaging
\end{keywords}

\section{Introduction}
The ImageNet Large Scale Visual Recognition Challenge by \cite{deng2009imagenet} provides more than one million labeled images of 1,000 object categories.
The accessibility of a huge amount of well-annotated image data in computer vision rekindles deep convolutional neural networks (CNNs) as a premier learning tool to solve the visual object class recognition tasks, as shown by \cite{krizhevsky2012imagenet, simonyan2014very, Szegedy2014Going}.
Deep CNNs can perform significantly better than traditional shallow learning methods, but usually requires much more training data as was shown by \cite{krizhevsky2012imagenet,russakovsky2014imagenet}.
In the medical domain, however, there are no similar large-scale labeled image datasets available.
On the other hand, large collections of radiology images and reports are stored in many modern hospitals' Picture Archiving and Communication Systems (PACS).
The invaluable semantic diagnostic knowledge inhabiting the mapping between hundreds of thousands of clinician-created high quality text reports and linked image volumes remains largely unexplored.
One of our primary goals is to extract and associate radiology images with clinically semantic labels via interleaved text/image data mining and deep learning on a large-scale PACS database ($\sim$780K imaging examinations). 
To the best of our knowledge, this is the first reported work performing automated mining and prediction on a hospital PACS database at a very large scale.

The Radiology reports are text documents describing patient history, symptoms, image observations and impressions written by board-certified radiologists.
However, the reports do not contain specific image labels to be trained by a machine learning algorithm.
Building the ImageNet database (\cite{deng2009imagenet}) was mainly a manual process: harvesting images returned from Google image search engine according to the WordNet (\cite{miller1995wordnet}) ontology hierarchy and pruning falsely tagged images using crowd-sourcing such as Amazon Mechanical Turk (AMT).
This does not meet our data collection and labeling needs due to the demanding difficulties of medical annotation tasks and the data privacy reasons.
Thus we first propose to mine categorical semantic labels using non-parametric topic modeling method -- latent Dirichlet Allocation (LDA) by \cite{blei2003latent} -- to provide a semantic interpretation of a patient image in three levels.
While this provides a first-level interpretation of a patient image, labeling based on categorization can be nonspecific.
To alleviate the issue of non-specificity, we further mined specific disease words in the reports mentioning the images.
Feed-forward CNNs were then used to train and predict the presence/absence of the specific disease categories.

Our work has been inspired by the works of \cite{deng2009imagenet,russakovsky2014imagenet} building very large-scale image databases and the works establishing semantic connections of texts and images by \cite{berg2013babytalk}.
Please note, that there has not yet been much comparable development on large-scale medical imaging interpretation.
\cite{berg2013babytalk} have spearheaded the efforts of learning the semantic connections between image contents and the sentences describing them, such as image captions.
Detecting objects of interest, attributes and prepositions and applying contextual regularization with a conditional random field (CRF) is a feasible approach as shown by \cite{berg2013babytalk}, and many useful tools for image annotation using it are available in computer vision.

Both deep feed-forward CNNs of \cite{krizhevsky2012imagenet,simonyan2014very} and recurrent neural networks of \cite{mikolov2013efficient,mikolov2013distributed} were used to model image and text.
Also, the CNN parameters pre-trained on ImageNet were used to initialize CNNs and to be adopted for medical image analysis.
We show the benefit of this transfer learning and domain adaptation in Section \ref{sec:transfer_learning}.
The fact that deep learning requires no hand-crafted image features is very desirable since significant adaption would be needed to apply conventional image features, e.g., HOG, SIFT to learn the wide variety of medical images.
The large-scale datasets of extracted key images and their categorization, vector labels, and describing sentences can be harnessed to alleviate deep learning's ``data-hungry'' challenge in the medical domain. We will make our code and trained deep text/image models publicly available upon acceptance.

\subsection{Related Work}
The ImageCLEF medical image annotation tasks of 2005-2007 by \cite{Deselaers2008Deformations} have 9,000 training and 1,000 testing two-dimensional images, converted to $32 \times 32$ pixel thumbnails with $57$ labels.
Local image descriptors and intensity histograms are used as a bag-of-features approach in that work for this scene-recognition-like problem. 
Unsupervised LDA based matching from lung disease words (e.g., fibrosis, emphysema) to two-dimensional image blocks from axial CT chest scans of 24 patients is studied by \cite{Carrivick2005Unsupervised}.
The works of \cite{Barnard2003Matching,Blei2003Modeling} using generative models of combining words and images under a very limited word/image vocabulary has also motivated this study.


The most related works are by \cite{socher2013zero,Frome2013devise} where they first map words into vector space using recurrent neural networks and then project images into the label-associated word-vector embeddings by minimizing the $L_2$ (\cite{socher2013zero}) or hinge rank losses (\cite{Frome2013devise}) between the visual and label manifolds.
The language model is trained on the texts of Wikipedia and tested on label-associated images from the CIFAR (\cite{krizhevsky2009learning,socher2013zero}) and ImageNet datasets (\cite{deng2009imagenet,Frome2013devise}).
Image-to-language correspondence was learned from ImageNet dataset and reasonably high quality image description datasets (Pascal1K (\cite{rashtchian2010collecting}), Flickr8K (\cite{hodosh2013framing}), Flickr30K (\cite{young2014image})) by \cite{karpathy2014deep}, where such caption datasets are not available in the medical domain.

Graphical models have been employed to predict image attributes by \cite{lampert2014attribute,Scheirer2012attribute}, or to describe images by \cite{berg2013babytalk} using manually annotated datasets.
Automatic label mining on large, unlabeled datasets is presented by \cite{ordonez2011im2text,jaderberg2014deep}, however the variety of the label-space is limited to image text annotations.
We analyze and mine the medical image semantics on both document and sentence levels, and deep CNNs of \cite{jaderberg2014deep,simonyan2014very} are adapted to learn them from image contents.

\section{Data}
\label{sec:data}

\begin{figure}[t]
\begin{center}
   \includegraphics[width=1.0\linewidth]{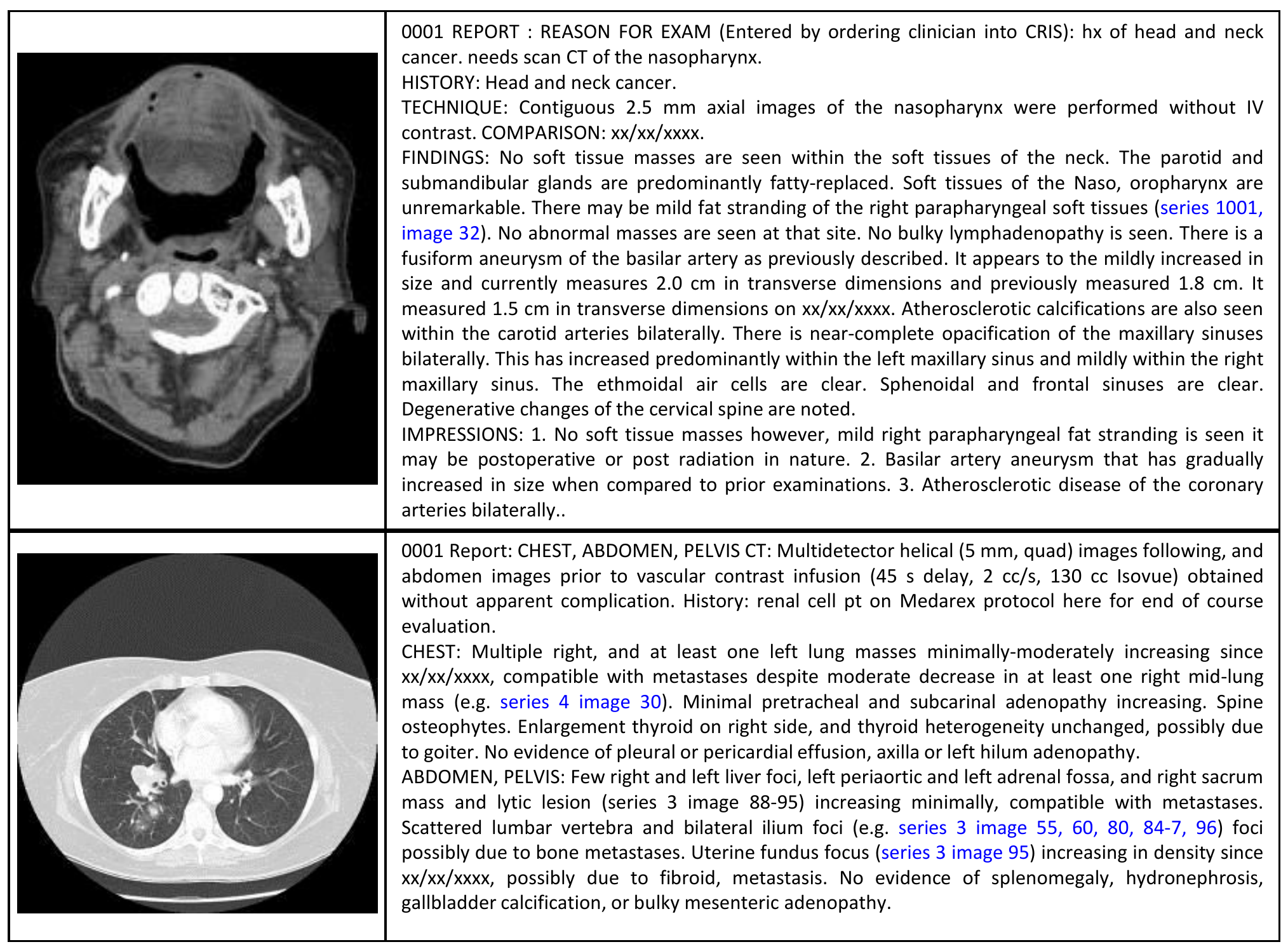}
\end{center}
   \caption{Two examples of radiology reports and the referenced ``key images'' (providing a visual reference to pathologies or other notable findings).}
\label{fig:reports_images_01}
\end{figure}

To gain the most comprehensive interpretation of diagnostic semantics, we use all available radiology reports of around 780K imaging examinations, stored in the PACS of National Institutes of Health Clinical Center since the year 2000.
Around $216K$ key two-dimensional image slices are studied here, instead of using all three-dimensional image volumes.
Within three-dimensional patient scans, most of the imaging information represented is normal anatomy, therefore they are often not the focus of the radiology reports.
The two-dimensional ``key images'' referenced (Figure \ref{fig:reports_images_01}) by radiologists manually during radiology report writing provide a visual reference to pathologies or other notable findings.
Therefore, the two-dimensional key images are more correlated with the diagnostic semantics in the reports than the whole three-dimensional scans, but not all reports have referenced key images ($215,786$ images from about $61,845$ unique patients).
Table \ref{tab:ris_stats} provides some statistics of the extracted database, and Table \ref{tab:frequently_occuring_words_ris_reports} shows examples of the most frequently occurring words in radiology reports collected.
Leveraging our deep learning models exploited in this paper will make it possible to automatically select key images from three-dimensional patient scans to avoid mis-referencing.

Finding and extracting key images from radiology reports is done by natural language processing (NLP), i.e, finding a sentence mentioning a referenced image.
For example, ``\textsl{There may be mild fat stranding of the right parapharyngeal soft tissues (series 1001, image 32)}'' is listed in Figure \ref{fig:reports_images_01}.
The NLP steps are sentence tokenization, word/number matching and stemming, 
and rule-based information extraction (e.g., translating ``image 1013-78'' to ``images 1013-1078'').
Software package of \cite{bird2009natural} was used for basic NLP pipelines.
A total of $\sim$187K images could be retrieved and matched this way, whereas the rest of $\sim$28K key images were extracted according to their reference accession numbers in PACS.

\begin{table}
\begin{center}
\resizebox{.7\linewidth}{!}{
\begin{tabular}{|l|c||c|c||c|c|}
\hline
\multicolumn{2}{|c||}{total number of} & \multicolumn{2}{c||}{\# words in documents} & \multicolumn{2}{c|}{\# image modalities} \\
\hline\hline
\# documents  & $\sim$780k      & \texttt{mean}   & 131.13 & \textmd{CT}     & $\sim$169k \\
\hline
\# images     & $\sim$216k      & \texttt{std}    & 95.72  & \textmd{MR}     & $\sim$46k  \\
\hline
\# words      & $\sim$1 billion & \texttt{max}    & 1502   & \textmd{PET}    & 67         \\
\hline
\# vocabulary & $\sim$29k       & \texttt{min}    & 2      & \textmd{others} & 34       \\
\hline
\end{tabular}
}
\end{center}
\caption{Some statistics of the dataset. ``Others'' include computed radiography, and ultrasound.}
\label{tab:ris_stats}
\end{table}

\begin{table}
\begin{center}
\resizebox{.7\linewidth}{!}{
\begin{tabular}{|c|c||c|c||c|c||c|c|}
\hline
right      & 937k & images & 312k & contrast & 260k & unremarkable & 195k \\
\hline
left       & 870k & seen   & 299k & axial    & 253k & lower        & 195k \\
\hline
impression & 421k & mass   & 296k & lung     & 243k & upper        & 192k \\
\hline
evidence   & 352k & normal & 278k & bone     & 219k & lesion       & 180k \\
\hline
findings   & 340k & small  & 275k & chest    & 208k & lobe         & 174k \\
\hline
CT         & 312k & noted  & 263k & MRI      & 204k & pleural      & 172k \\
\hline
\end{tabular}
}
\end{center}
\caption{Examples of the most frequently occurring words in the radiology report documents.}
\label{tab:frequently_occuring_words_ris_reports}
\end{table}

\section{Document Topic Learning with Latent Dirichlet Allocation}
\label{sec:document_to_topic_and_image}

We propose to mine image categorization labels using non-parametric topic-modeling algorithm of \cite{blei2003latent} on the \texttildelow 780K radiology text reports in PACS.
Unlike the images of ImageNet (\cite{deng2009imagenet,russakovsky2014imagenet}) which often have a dominant object appearing in the center, our key images are mostly CT and MRI slices showing several organs usually with pathologies.
There are high amounts of intrinsic ambiguity in defining and assigning a semantic label set to images, even for experienced clinicians.
Our hypothesis is that the large collection of sub-million radiology reports statistically defines the categories meaningful for topic-mining and visual correspondence learning for these topics.

Latent Dirichlet Allocation (LDA) was originally proposed by \cite{blei2003latent} to find latent topic models for a collection of text documents such as newspaper articles.
There are some other popular methods for document topic modeling, such as Probabilistic Latent Semantic Analysis (pLSA) by \cite{hofmann1999probabilistic} and Non-negative Matrix Factorization (NMF) by \cite{lee1999learning}.
We choose LDA for extracting latent topic labels among radiology report documents, because LDA is shown to be more flexible yet learns more coherent topics over large sets of documents, as was studied by \cite{stevens2012exploring}.
Furthermore, pLSA can be regarded as a special case of LDA (\cite{girolami2003equivalence}) and NMF as a semi-equivalent model of pLSA (\cite{gaussier2005relation,ding2006nonnegative}).

LDA offers a hierarchy of extracted topics and the number of topics can be chosen by evaluating each model's {\em perplexity score} (Equation \ref{eq:lda_perplexity}), which is a common way to measure how well a probabilistic model generalizes by evaluating the log-likelihood of the model on a held-out test set.
For an unseen document set $D_\text{test}$, the perplexity score is defined as in Equation \ref{eq:lda_perplexity}, where $M$ is the number of documents in the test set, $\textbf{w}_d$ the words in the unseen document $d$, $N_d$ the number of words in document $d$, with $\mathbf{\Phi}$ the topic matrix, and $\alpha$ the hyper-parameter for topic distribution of the documents.

\begin{equation}
   perplexity(D_{test}) = \exp \left\{ - \frac{\sum^{M}_{d=1}\log p(\mathbf{w}_d|\mathbf{\Phi},\alpha)}{\sum^{M}_{d=1}N_d}  \right\}
\label{eq:lda_perplexity}
\end{equation}

\noindent
A lower perplexity score generally implies a better fit of the model for a given document set (\cite{blei2003latent}).

\begin{figure}[t]
\begin{center}
   \includegraphics[width=.68\linewidth]{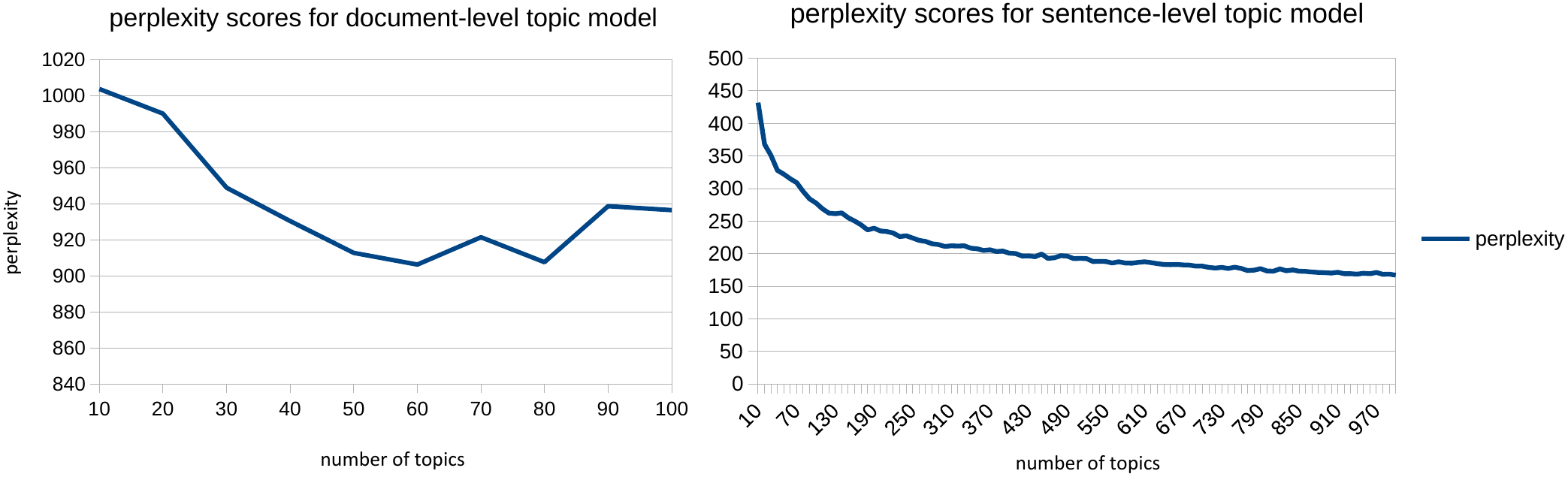}
\end{center}
   \caption{Perplexity scores for document-/sentence- level topic models. Number of topics with low perplexity score is selected as the optimal (80 for document-level, 1000 for sentence-level).}
\label{fig:lda_topic_perplexity}
\end{figure}

\begin{figure}[t]
\begin{center}
   \includegraphics[width=1.0\linewidth]{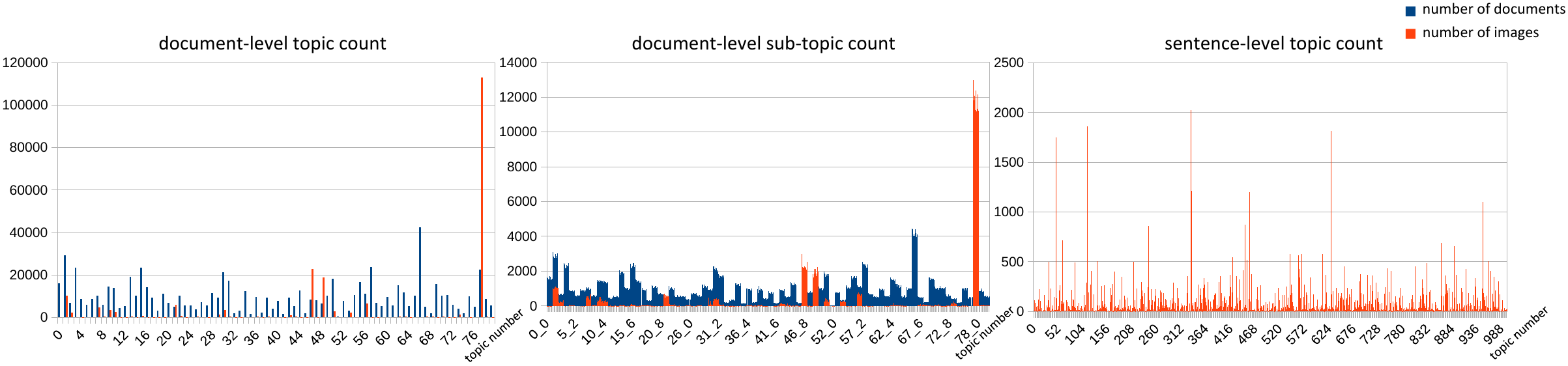}
\end{center}
   \caption{Distribution of documents and images for document-level topic, document-level sub-topic, and sentence-level topic. Sixth sub-topic of document topic 41 is noted as \texttt{$5\_5$}.}
\label{fig:lda_topic_distribution}
\end{figure}

Based on the perplexity score evaluated on 80\% of the total documents used for training and 20\% used for testing, the number of topics chosen is 80 for the document-level model using perplexity scores for model selection (Figure \ref{fig:lda_topic_perplexity}).
Although the document distribution in the topic space is approximately balanced, the distribution of image counts for the topics is more unbalanced (Figure \ref{fig:lda_topic_distribution}).
Specifically, topic $\#77$ (non-primary metastasis spreading across a variety of body parts) contains nearly half of the $~$216K key images.
To address this data bias, sub-topics are obtained for each of the first document-level topics, resulting in 800 topics, where the number of the sub-topics is also chosen based on the average perplexity scores evaluated on each document-level topic.
Lastly, to compare the method of using the whole report with using only the sentence directly describing the key images for latent topic mining, a sentence-level LDA topics are obtained based on three sentences only: the sentence mentioning the key-image (Figure \ref{fig:reports_images_01}) and its adjacent sentences as proximal context.
The perplexity scores keep decreasing with an increasing number of topics; we choose the topic count to be 1000 as the rate of the perplexity score decrease is very small beyond that point (Figure \ref{fig:lda_topic_perplexity}).

\begin{figure}[t]
\begin{center}
   \includegraphics[width=1.0\linewidth]{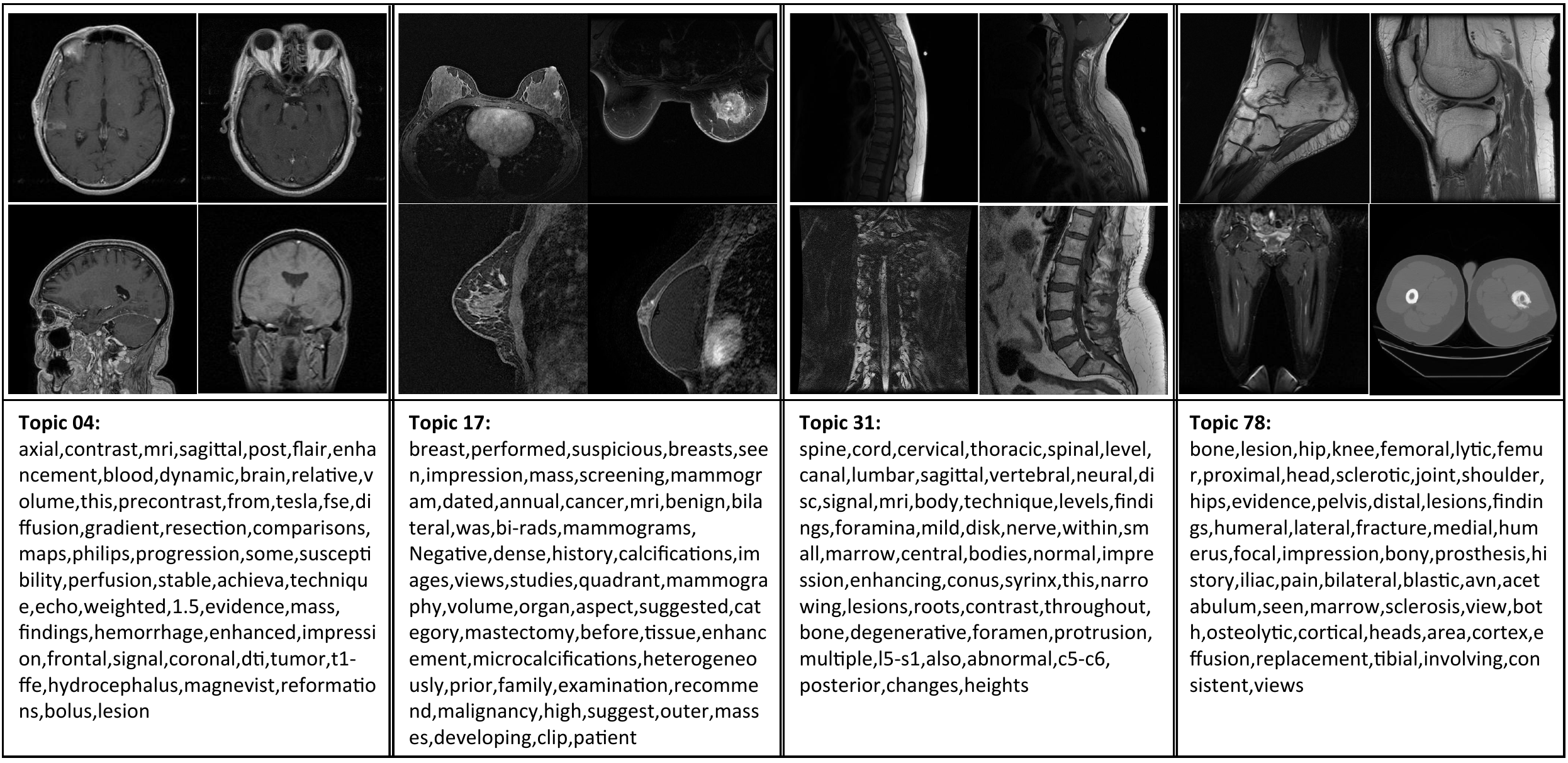}
\end{center}
   \caption{Examples of LDA generated document-level topics with corresponding images and key words. Topic $\#4$ is MRI of brain tumor; topic $\#17$: breast imaging; topic $\#31$: degenerative spine disc disease; and topic $\#78$: bone metastases. These are verified by a radiologist.}
\label{fig:lda_document_level_topic_distribution_example_01}
\end{figure}

We observe that LDA-generated image categorization labels are valid, demonstrating good semantic coherence among clinician observers. 
The lists of key words and sampled images per topic label are subjected to board-certified radiologist's review and validation.
Some examples of document-level topics with their corresponding images and topic key words are shown in Figure \ref{fig:lda_document_level_topic_distribution_example_01}.
Based on radiologists' review, our LDA topics discover semantics at different levels. 
There are 73 low-level concepts for example, pathology examination of certain body regions and organs: topic $\#47$ - sinus diseases; $\#2$ - lesions of solid abdominal organs, primarily kidney; $\#10$ - pulmonary diseases; $\#13$ - brain MRI; $\#19$ - renal diseases on mixed imaging modalities; $\#36$ - brain tumors.
There are 7 mid- to high-level concepts, such as: topic $\#77$ - non-primary metastasis spreading across a variety of body parts; topic $\#79$ - cases with high diagnosis uncertainty or equivocation; $\#72$ - indeterminate lesions; $\#74$ - instrumentation artifacts limiting interpretation.
Low-level topic images tend to be visually more coherent than the higher-level topic images.

\begin{figure}[t]
\begin{center}
   \includegraphics[width=1.0\linewidth]{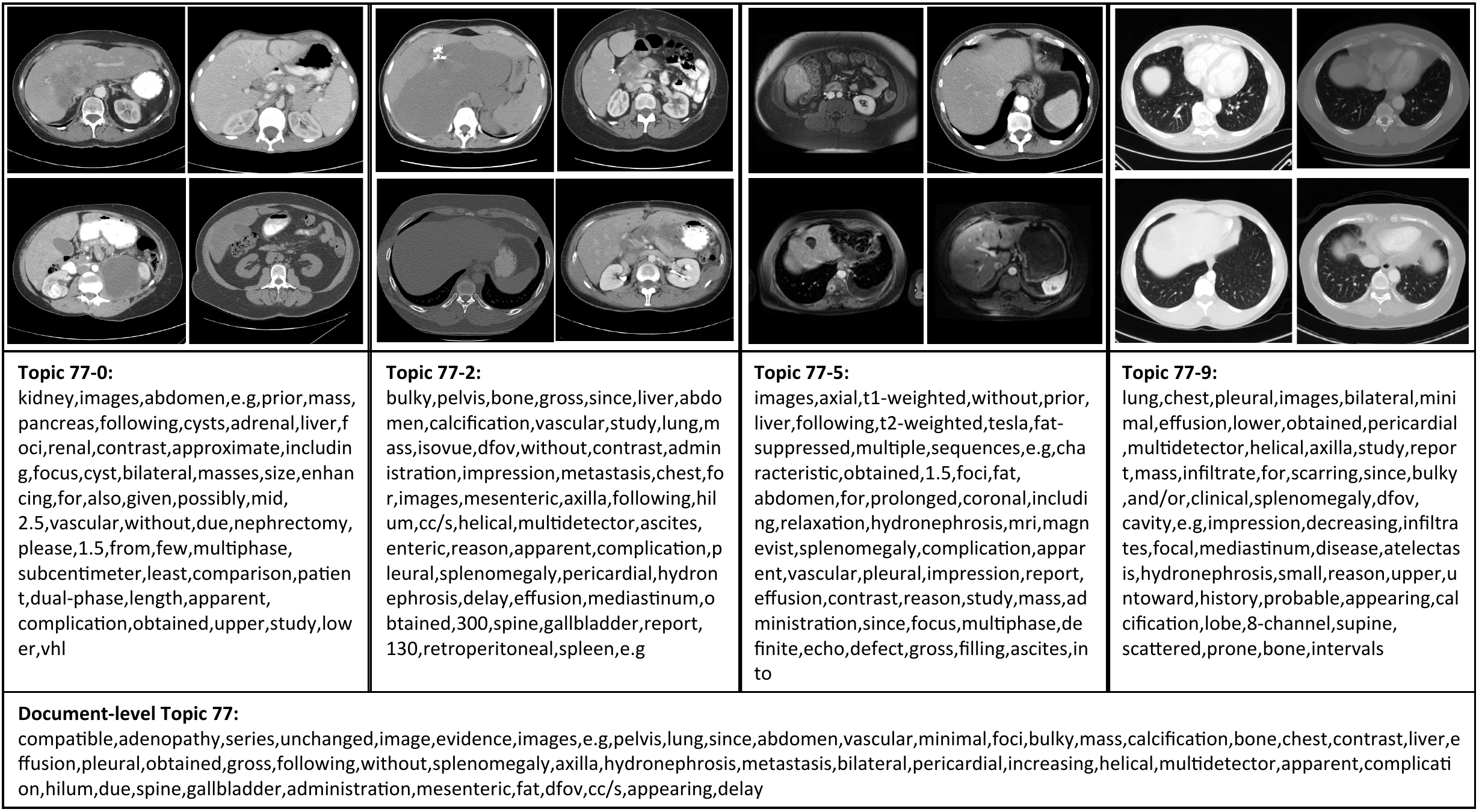}
\end{center}
  \caption{
  Examples of some sub-topics of document-level topic \#77, with corresponding images and topic key-words. 
  The key-words and the images for the document-level topic (\#77) indicates metastatic disease. The key-words for topic \#77 are: [\footnotesize{\textmd{abdomen,pelvis,chest,contrast,performed,oral,was,present,masses,stable,intravenous,adenopathy,\newline liver,retroperitoneal,comparison,administration,scans,130,small,parenchymal,mediastinal,dated,\newline after,which,evidence,were,pulmonary,made,adrenal,prior,pelvic,without,cysts,spleen,mass,disease,\newline multiple,isovue-300,obtained,areas,consistent,nodules,changes,pleural,lesions,following,abdominal,\newline that,hilar,axillary}}].}
\label{fig:lda_document_level_h2_topic_distribution_example_01}
\end{figure}

High-level topics may be analogous to the high-level visual concepts in natural images as was studied by \cite{Kiapour2014Hipster,Ordonez2014High}.
About half of the key images are associated with topic $\#77$, implying that the clinicians' image referencing behavior patterns heavily focuses on metastatic patients.
Sub-topics of document-level topic $\#77$ are sub-categories of metastatic disease, for example: $\#77$-$0$ - abdominal mass; $\#77$-$2$ - bulky tumor; $\#77$-$4$ - multifocal metastatic disease; $\#77$-$9$ - liver tumor.
Meanwhile, some of the sub-topics of document-level $\#77$ do not seem very focused.
Many of the sentence-level topics have valid semantics too, e.g. `renal imaging', `musculoskeletal imaging', `chest port catheter', `chest imaging with disease or pathology', and `degenerative disease in bone'.

We also obtained LDA topics on the reports having associated images only, resulting in 20 topics according to perplexity score.
However, these did not add any more meaningful semantics in addition to the already obtained topics in three levels, so that we did not include the topics.
For more details and the image-topic associations, refer to Figures \ref{fig:lda_document_level_topic_distribution_example_01}, \ref{fig:lda_document_level_h2_topic_distribution_example_01}, and the supplementary material.
Even though LDA labels are computed with text information only, we next investigate the plausibility of mapping images to the topic labels of different levels via deep CNN models.

\section{Image to Document Topic Mapping with Deep Convolutional Neural Networks}
\label{sec:image_to_topic}

\begin{figure}[t]
\begin{center}
   \includegraphics[width=1.0\linewidth]{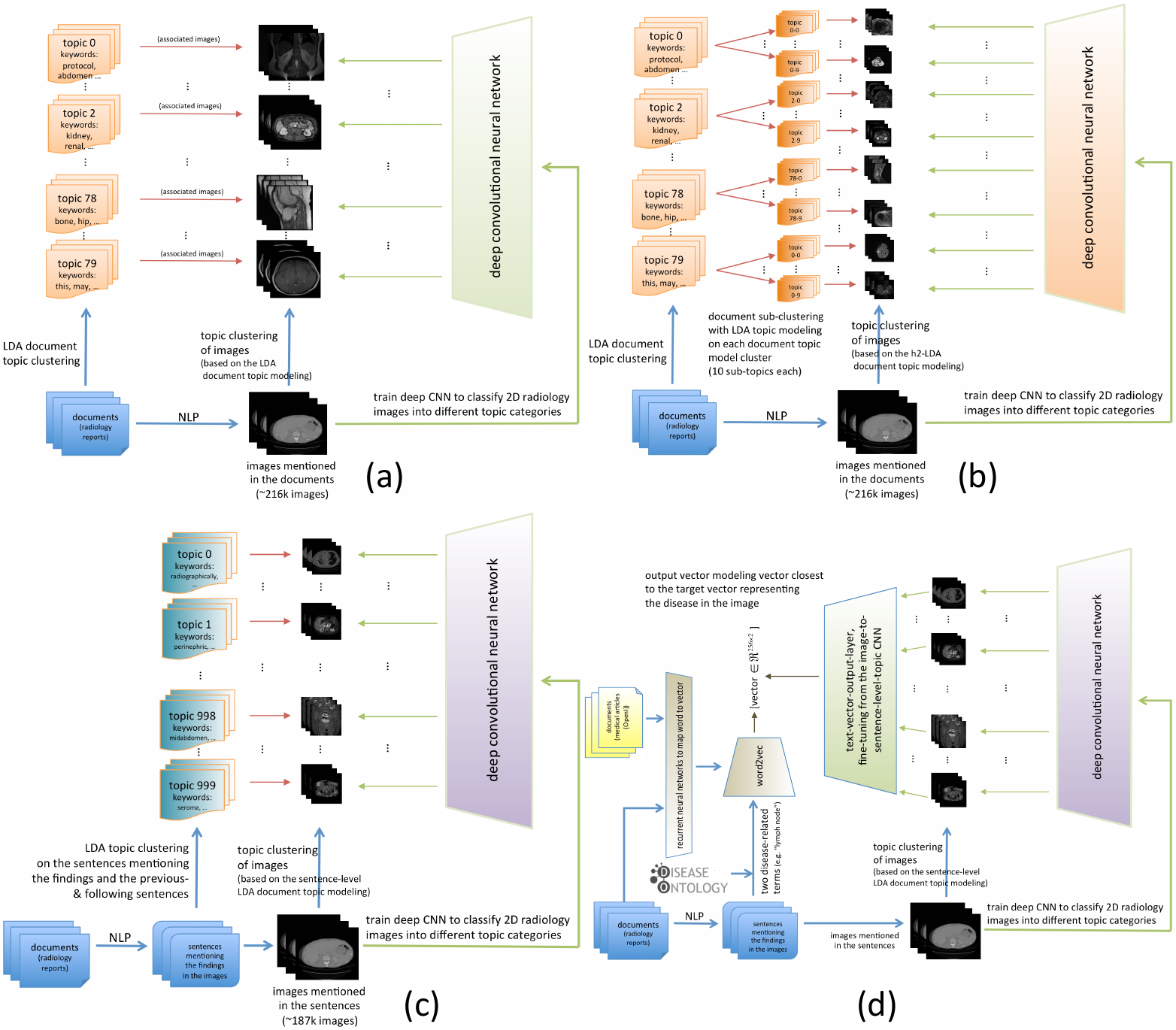}
\end{center}
   \caption{Systematic diagrams for training CNNs to learn to classify images into (a) document-level topics (b) document-level sub-topics, (c) sentence-level topics. A systematic diagram for image-to-word model (in Section \ref{sec:bi-gram-cnn}) is shown in (d).}
\label{fig:sys-diagrams-h123}
\end{figure}

For each level of topics discussed in Section \ref{sec:document_to_topic_and_image}, we train deep CNNs to map the images into document categories using the Caffe framework of \cite{jia2014caffe}.
We split our whole key image dataset as follows: 85\% used as the training dataset, 5\% as the cross-validation (CV), and 10\% as the test dataset.
If a topic has too few images to be divided into training/CV/test for deep CNN learning, then that topic is neglected for the CNN training.
These cases are normally topics of rare imaging protocols, for example: topic $\#5$ - Abdominal ultrasound; topic $\#28$\& $\#49$ - DEXA scans of different usages.
In total, 60 topics were used for the document-level topic mapping, 385 for the document-level sub-topic mapping, and 717 for the sentence-level mapping.
Systematic diagrams showing how each level of semantic topics are learned, assigned to images, and trained to map from images to topics are shown in Figure \ref{fig:sys-diagrams-h123}.

\subsection{Implementation}
All our CNN network settings are similar or the same as the ImageNet Challenge ``AlexNet'' (\cite{krizhevsky2012imagenet}), and ``VGG-16 \& 19'' (\cite{simonyan2014very}) models.
For ``AlexNet'' we used the Caffe reference network of \cite{jia2014caffe}, which is a slight modification to the ``AlexNet'' by \cite{krizhevsky2012imagenet}.
The AlexNet model by\cite{krizhevsky2012imagenet} has about 60 million parameters (650,000 neurons) and consists of five convolutional layers (1st, 2nd and 5th followed by max-pooling layers), and three fully-connected (FC) layers with a final classification layer.
The VGG variations of CNN models by \cite{simonyan2014very} are significantly deeper by having 16$\sim$19 convolutional layers and 133$\sim$144 million parameters.
The top-1 error rates on ImageNet dataset of these models are AlexNet:15.3\% (\cite{krizhevsky2012imagenet}); VGG-16:7.4\%; and VGG-19:7.3\% (\cite{simonyan2014very}), respectively.

For image to topic mapping, we change the numbers of output nodes in the last softmax classification layer, i.e., 60, 385 and 717 for the document-level, document-level sub-topics, and sentence-level respectively.
The networks for first-level semantic labels are fine-tuned from the pre-trained ImageNet models, where the networks for the lower-level semantic labels are fine-tuned from the models of the higher-level semantic labels.

\subsection{Transfer Learning and Domain Adaptation}
\label{sec:transfer_learning}
\begin{figure}[t]
\begin{center}
   \includegraphics[width=1.0\linewidth]{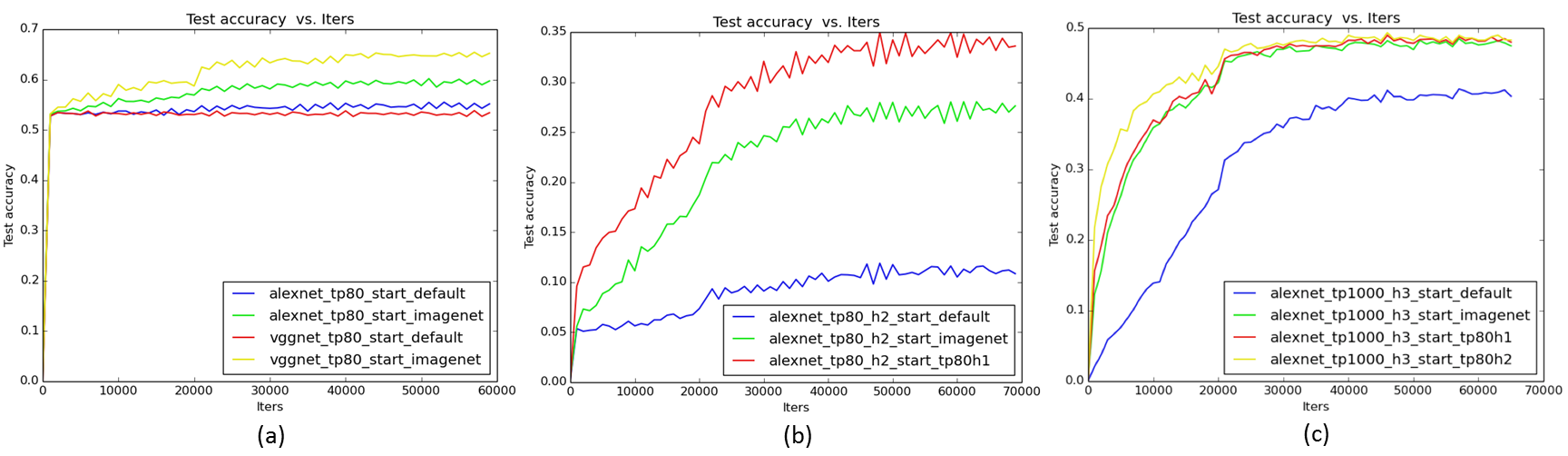}
\end{center}
   \caption{Traces of classification accuracies during training, showing benefits of using ImageNet dataset as pre-training for this task with medical images and improvements of fine-tuning from CNN neural networks of similar tasks (e.g., from document-level (h1) CNN model to document-level sub-topic (h2) CNN model). (a) Image-to-document-level-topic (h1) classification, (b) image-to-document-level-sub-topic (h2) classification, and (c) image-to-sentence-level-topic (h3) classification.}
\label{fig:train_accs_h123}
\end{figure}

We found that transfer learning from the ImageNet pre-trained CNN parameters on natural images to our medical image modalities (mostly CT, MRI) significantly helps the image classification performance. 
Additionally, transfer learning from a CNN trained for a more related task (e.g. from CNN trained on image-to-document-level-topic models to train CNN for image-to-document-level-sub-topic model) was found to be more effective than from a CNN trained for a less related task (e.g. from CNN trained on ImageNet to train CNN for image-to-document-level-sub-topic model).
Examples of classification accuracy traces during training using CNNs from random initialization, transfer learning from CNN trained on ImageNet, and transfer learning from higher level image-to-topic model to lower level image-to-topic models are shown in Figure \ref{fig:train_accs_h123}.
Similar findings that deep CNN features can be generalized across different image modalities have been reported by \cite{Gupta2014rgbd,Gupta2013Natural}, but are empirically verified with only much smaller datasets than ours.
Our key image dataset is about one fifth the size of ImageNet (\cite{russakovsky2014imagenet}) and is the largest annotated medical image dataset to date.

From Figure \ref{fig:train_accs_h123} we can see that: (1) CNN testing accuracy increases from $\sim$0\% to 50+\% quickly in about 1600 iterations due to the unbalanced data distribution among classes in document-level; (2) A more complex, deeper CNN model (VGG-Net) performs better than the model which already is a good benchmark (AlexNet), but only when starting from a good initialization (i.e., pre-training via ImageNet models); (3) Fine-tuning from a more closely related task CNN model is even better than fine-tuning from less related task model (alexnet\_tp80\_h2\_start\_tp80h1 $>$ alexnet\_tp80\_h2\_start\_imagenet).

With these findings, we train our CNN models with transfer-learning by default for the remaining parts of our study.
All the CNN layers except the newly modified ones are initialized with the weights of a previously trained related model, and trained with a new task with low learning rate of 0.001.
The modified layers with new number of classes are initialized randomly, and their learning rates are set with higher learning rate of 0.01. 
All the key images are resampled to a spatial resolution of $256\times256$ pixels. 
Then we follow the approach of \cite{simonyan2014very} to crop the input images from $256\times256$ to $227\times227$ for training.

\subsection{Classification Results and Discussion}
We would expect that the level of difficulties for learning and classifying the images into the LDA-induced topics will be different for each semantic level.
Low-level semantic classes can have key images of axial/sagittal/coronal slices with position variations and across MRI/CT modalities.
Mid- to high-level concepts all demonstrate much larger within-class variations in their visual appearance since they are diseases occurring within different organs and are only coherent at high level semantics.
Table \ref{tab:image_to_topic_mapping_CNN_acc} provides the top-1 and top-5 testing in classification accuracies for each level of topic models using AlexNet (\cite{krizhevsky2012imagenet}), and VGG-16\&19 (\cite{simonyan2014very}) based deep CNN models.

\begin{figure}[t]
\begin{center}
   \includegraphics[width=1.0\linewidth]{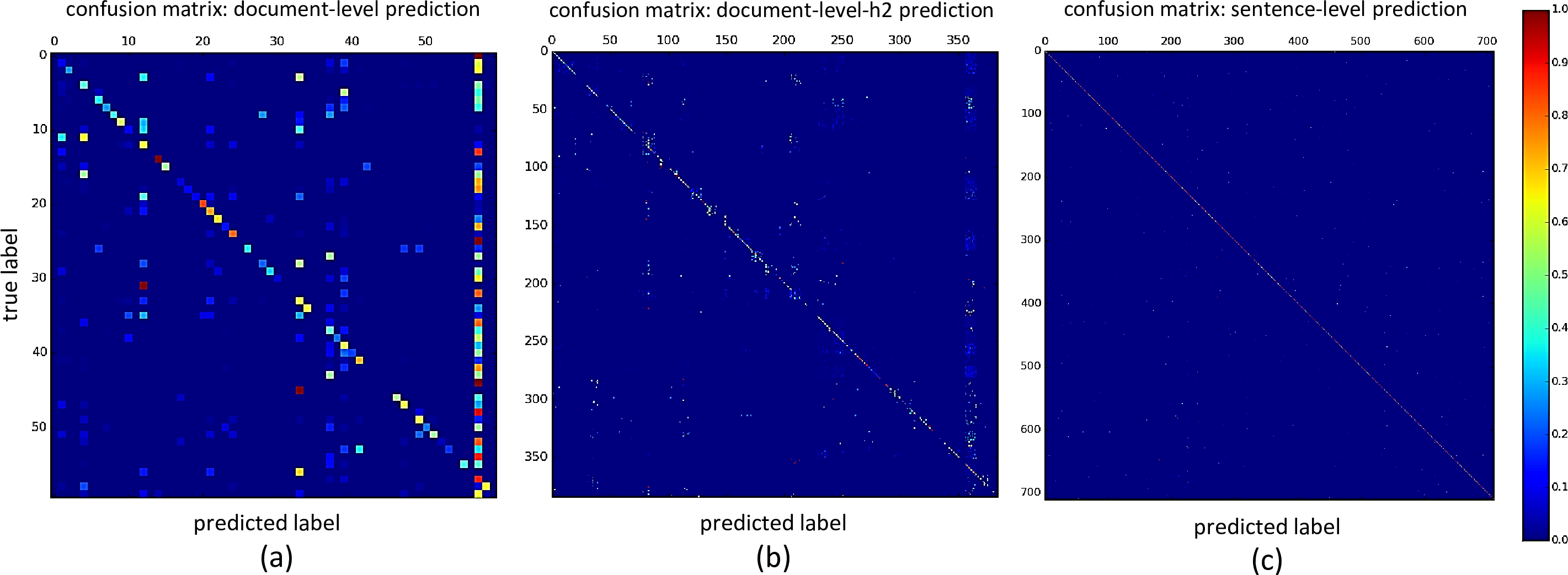}
\end{center}
   \caption{\small Confusion matrices of (a) document-level topic, (b) document-level sub-topic (document-level-h2), and (c) sentence- level classification \cite{simonyan2014very} ((b) and (c) can be viewed best in electronic version of this document).}
\label{fig:conf_matrices}
\end{figure}

All top-5 accuracy scores are significantly higher than top-1 values, e.g. increasing from $0.658$ to $0.946$ using VGG-19, or $0.607$ to $0.929$ via AlexNet in document-level.
This indicates that the classification errors or fusions are not uniformly distributed among other false classes.
Latent ``blocky subspace of classes'' may exist in our discovered label space, where several topic classes form a tightly correlated subgroup.
The confusion matrices in Figure \ref{fig:conf_matrices} verify this finding.

It is shown that the deeper models (VGG-16\&19) perform consistently better than the shallower 8-layer model (AlexNet) in classification accuracy, especially for document-level sub-topics.
While the images of some topic categories and some body parts are easily distinguishable as shown in Figure \ref{fig:lda_document_level_topic_distribution_example_01}, the visual differences in abdominal parts are rather subtle as in Figure \ref{fig:lda_document_level_h2_topic_distribution_example_01}.
Distinguishing the subtleties and high-level concept categories in the images could benefit from a more complex model so that the model can handle these subtleties. 

It is also noticeable that deeper models require significantly more computational resource and time to train than the shallower model.
Table \ref{tab:image_to_topic_mapping_CNN_time_mem} shows the memory consumption and time required to train the CNN models for the image-to-sentence-level-topic model with up to 70,000 iterations using the NVidia Tesla K40 GPU.
However, comparing VGG-16 and VGG19, three additional convolutional layers seem to have contributed to raise the top-5 accuracies by a small amount ($\sim$2\%), which is coherent with the results reported by \cite{simonyan2014very} for object recognition task on the ImageNet dataset.

Compared with the ImageNet 2014 results, top-1 error rates are moderately higher ($34\%$ versus $30\%$) and top-5 test errors ($6\%\sim8\%$) are comparable.
In summary, our quantitative results are very encouraging, but there also exist some uncertainties in annotations because labels stem from an unsupervised learning algorithm.
Multi-level semantic concepts show good image learnability by deep CNN models which sheds light on the feasibility of automatically parsing very large-scale radiology image databases.

\begin{table}
\begin{center}
\resizebox{.73\linewidth}{!}{
\begin{tabular}{|c||c|c||c|c||c|c|}
\hline
 & \multicolumn{2}{c||}{AlexNet 8-layers} & \multicolumn{2}{c||}{VGG 16-layers} & \multicolumn{2}{c|}{VGG 19-layers}  \\ \cline{2-7}
                  & top-1   & top-5 & top-1 & top-5 & top-1 & top-5  \\
\hline \hline

document-level    & 0.61  & 0.93  & 0.66 & 0.93  & 0.66  & 0.95  \\
\hline

document-level-h2 & 0.33  & 0.56 & 0.54 & 0.70 & 0.54  & 0.70  \\ \hline

sentence-level & 0.48 & 0.56 & 0.50 & 0.56 & 0.50 & 0.58  \\

\hline
\end{tabular}
}
\end{center}
\caption{Top-1, top-5 test classification accuracies for image to document-level topics, document-level sub-topics (document-level-h2) and sentence-level topics, using AlexNet (\cite{krizhevsky2012imagenet}), and VGG-16\&19 (\cite{simonyan2014very}) deep CNN models.} \label{tab:image_to_topic_mapping_CNN_acc}
\end{table}

\begin{table}[t]
\begin{center}
\resizebox{.65\linewidth}{!}{
\begin{tabular}{|c||c|c|c|c|}
\hline
               & AlexNet 8-layers & VGG 16-layers & VGG 19-layers \\
\hline\hline
time    & 4 hours 35 mins & 3 days 2 hours & 4 days 40 mins \\
\hline
memory  & $\sim$1.4 GBytes & $\sim$10 GBytes & $\sim$11 GBytes \\
\hline
\end{tabular}
}
\end{center}
\caption{Training times for the CNN models used to reach 70,000 iterations,and their memory consumption, using Caffe framework (\cite{jia2014caffe}) on NVidia Tesla K40 GPU.}
\label{tab:image_to_topic_mapping_CNN_time_mem}
\end{table}

\section{Generating Image-to-Text Description}
\label{sec:image_to_text}

The image-to-topic mapping in Section \ref{sec:image_to_topic} is a promising first step towards automated interpretation of medical images in large scale.
However, it is too expensive and time consuming for radiologists to examine all of the 1880 (80 + 800 + 1000) topics generated with their keywords and images.
In addition, key words in the topics can help to understand the semantic contents of a given image with more semantic meaning.
We therefore propose to generate relevant key-word text descriptions similar to \cite{berg2013babytalk}, using deep language/image CNN models.



\subsection{Word-to-Vector Modeling and Removing Word-Level Ambiguity}
\label{subsec:word2vec}

\begin{figure}[t]
\begin{center}
   \includegraphics[width=.75\linewidth]{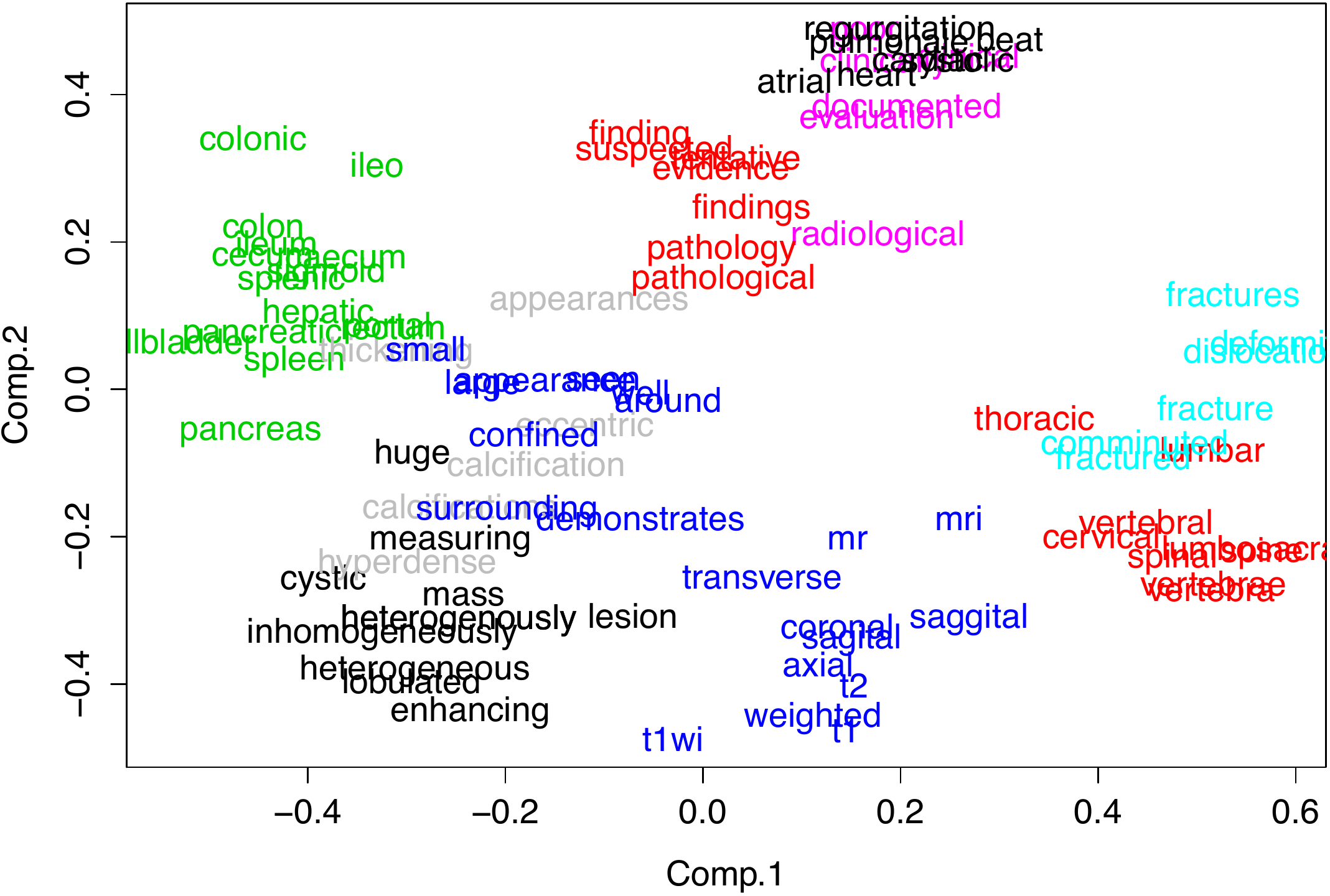}
\end{center}
   \caption{Example words embedded in the vector space using word-to-vector modeling (https://code.google.com/p/word2vec/) visualized on 2D space, showing (clinical) words with similar meanings are located nearby in the vector space (colors are used to highlight these in visualization).}
\label{fig:wordvec_plot_example_01}
\end{figure}

In radiology reports, there exist many recurring word morphisms in text identification, e.g., [mr, mri, t1-/t2-weighted] (natural language expressions for imaging modalities of magnetic resonance imaging (MRI)), [cyst, cystic, cysts], [tumor, tumour, tumors, metastasis, metastatic], etc.
We train a deep word-to-vector model of \cite{mikolov2013linguistic,mikolov2013distributed,mikolov2013efficient} to address this word-level labeling space ambiguity, while also transforming the words to vectors.
A total of $~$1.2 billion words from our radiology reports as well as from biomedical research articles obtained from OpenI (\cite{openi}ni: \url{http://openi.nlm.nih.gov}) are used.
Words with similar meaning are mapped or projected to closer locations in the vector space than dissimilar ones. 
An example visualization of the word vectors on a two-dimensional space using principal component analysis is shown in Figure \ref{fig:wordvec_plot_example_01}.

\begin{figure}[t]
\begin{center}
   \includegraphics[width=.8\linewidth]{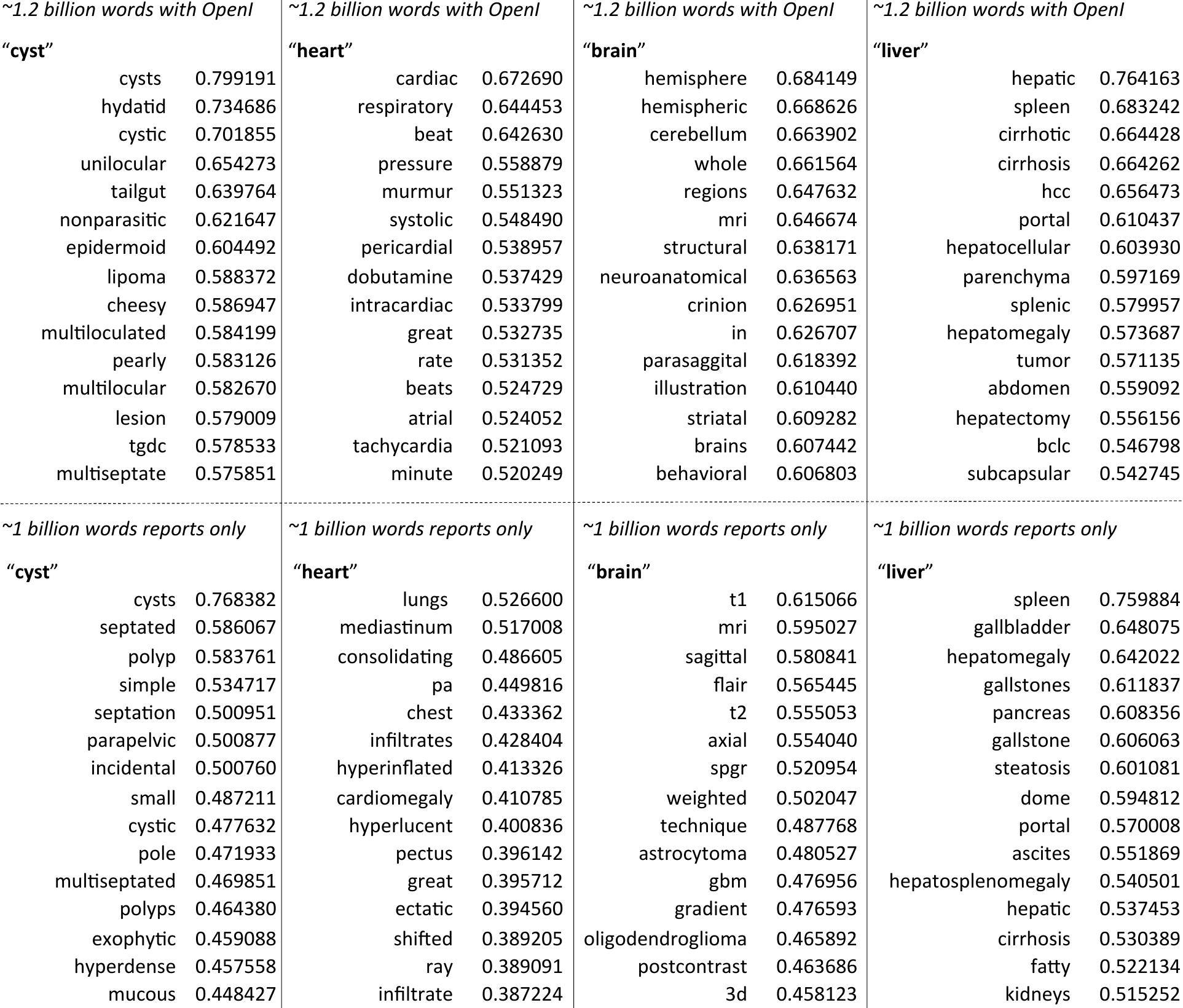}
\end{center}
   \caption{Word-to-vector models trained on a collection of biomedical research articles (from OpenI \cite{openi}) and radiology reports, and radiology reports only. Search words (with quotes) and their closest words in vector-space cosine similarity (higher the better) are listed in a descending order.} 
\label{fig:wordvec_openi_vs_reports_only}
\end{figure}

A skip-gram model of \cite{mikolov2013efficient,mikolov2013distributed} is employed with the mapping vector dimension of $\mathbb{R}^{256\times1}$ per word, trained using the {\em hierarchical softmax} cost function, sliding-window size of 10 and frequent words sub-sampled in frequency of 0.01.
It is found that combining an additional, more diverse set of related documents such as OpenI biomedical research articles, is helpful for the model to learn a better vector representation while keeping all the hyper-parameters the same.
Similar findings on unsupervised feature learning models, that robust faetures can be learned from a slightly noisy and diverse set of input, were reported by \cite{vincent2010stacked,vincent2008extracting,shin2013stacked}.
Some examples of query words and their corresponding closest words in terms of cosine similarity for the word-to-vector models (\cite{mikolov2013linguistic}), trained on radiology reports only (total of \texttildelow 1 billion words) and with additional OpenI articles (total of $~$1.2 billion words) are shown in Figure \ref{fig:wordvec_openi_vs_reports_only}.

\subsection{Image-to-Description Relation Mining and Matching}

The sentence referring to a key image and its adjacent sentences may contain a variety of words, but we are mostly interested in the disease-related terms which are highly correlated to diagnostic semantics.
To obtain only the disease-related terms, we exploit the human disease terms and their synonyms from the Disease-Ontology (DO; \cite{schriml2012disease}), a collection of 8,707 unique disease-related terms.
While the sentences referring to an image and their adjacent sentences have 50.08 words on average, the number of disease-related terms in the three consecutive sentences is $5.17$ on average with a standard deviation of $2.5$.
Therefore we chose to use bi-grams for the image descriptions, to achieve a good trade-off between the medium level complexity without neglecting too many text-image pairs.
Some statistics about the number of words in the documents are shown in Table \ref{tab:stats_word_lengths}.

\begin{table}
\begin{center}
\resizebox{.8\linewidth}{!}{
\begin{tabular}{|l||c|c|c|c|c|}
\hline
 \#words/sentence      & \texttt{mean} & \texttt{median} & \texttt{std} & \texttt{max} & \texttt{min} \\
\hline \hline
reports-wide & 11.70          & 9               & 8.97         & 1014         & 1 \\
\hline
image references       & 23.22 & 19 & 16.99 & 221 & 4 \\
\hline
image references, no stopwords no digits  & 13.46         & 11              & 9.94 & 143    & 2 \\
\hline
image references, disease terms only & 5.17  & 4               & 2.52         & 25           & 1 \\
\hline
\end{tabular}
}
\end{center}
\caption{Some statistics about number of words per sentence -- across the radiology reports (reports-wide), across the sentences identifying the key images and its two adjacent ones (image references) and these not counting stopwords and digits as well as counting disease related words only. }
\label{tab:stats_word_lengths}
\end{table}

\begin{figure}[t]
\begin{center}
   \includegraphics[width=.7\linewidth]{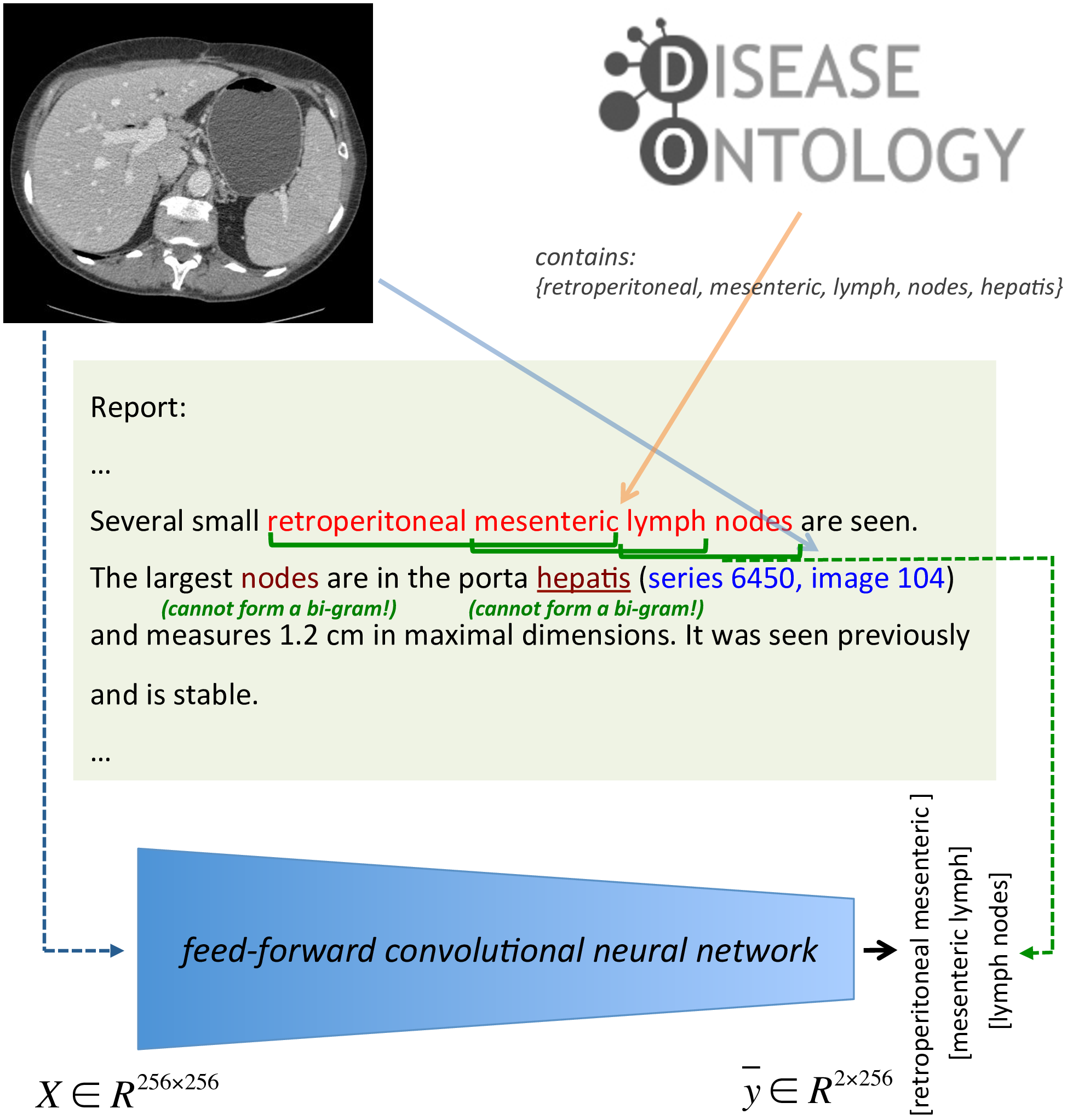}
\end{center}
   \caption{Illustration of how word sequences are learned for an image. Bi-grams are selected from the image's reference sentences containing disease-related terms from the disease ontology (DO; \cite{schriml2012disease}). Each bi-gram is converted to a vector of $\mathbf{Z} \in \mathbb{R}^{256\times2}$ to learn from an image. Image input vectors as $\{\mathbf{X}\in\mathbb{R}^{256\times256}$\} are learned through a CNN by minimizing the cross-entropy loss between the target vector and output vector. The words ``nodes'' and ``hepatis'' in the second line are DO terms but are ignored since they can not form a bi-gram.
   The DO logo is reproduced with permission from \url{http://disease-ontology.org}.
   }
\label{fig:bi_gram_nn}
\end{figure}

Bi-gram disease terms are extracted so that we can train a deep CNN model in Section \ref{sec:pipeline} to predict the vector/word- level image representation of $\mathbb{R}^{256\times2}$.
If multiple bi-grams can be extracted per image from the sentence referring the image and the two adjacent ones, the image is trained as many times as the number of bi-grams with different target vectors ($\mathbb{R}^{256\times2}$).
If a disease term cannot form a bi-gram, then the term is ignored, and the process is illustrated in Figure \ref{fig:bi_gram_nn}.
This is a challenging {\em weakly annotated learning} problem using referring sentences for labels.
The bi-grams of DO disease-related terms in the vector representation of $\mathbb{R}^{256\times2}$ are somewhat analogous to the work of \cite{berg2013babytalk} detecting multiple objects of interest and describing their spatial configurations in the image caption.
A deep regression CNN model is employed here, to map an image to a continuous output word-vector space from an image. 
The resulting bi-gram vector can be matched against a reference disease-related vocabulary in the word-vector space using cosine similarity. 

\subsection{Image-to-Words Deep CNN Regression}\label{sec:pipeline}
\label{sec:bi-gram-cnn}

It has been shown by \cite{sutskever2014sequence} that deep recurrent neural networks (RNN) can learn the language representation for machine translation.
To learn the image-to-text representation, we map the images to the vectors of word sequences describing the image.
This can be formulated as a regression CNN, replacing the softmax cost in Section \ref{sec:image_to_topic} with the cross-entropy cost function for the last output layer of VGG-19 CNN model (\cite{simonyan2014very}):

\begin{equation}
  E = -\frac{1}{n}\sum^N_{n=1}[g(\mathbf{z})_n \hat g(\bar{\mathbf{z}}_n) + (1-g(\mathbf{z}_n))\log(1-g(\hat{\mathbf{z}}_n))],
\end{equation}

\noindent
where $\mathbf{z}_n$ or $\hat{\mathbf{z}}_n$ is any uni-element of the target word vectors $\mathbf{Z}_n$ or optimized output vectors $\hat{\mathbf{Z}}_n$, $g(x)$ is the sigmoid function ($g(x) = 1/(1+e^{x})$), and $n$ is the number of samples in the database.

We adopt the CNN model of \cite{simonyan2014very} for the image-to-text representation since it works consistently better than the other relatively simpler model of \cite{krizhevsky2012imagenet} in our image-to-topic mapping tasks.
We fine-tune the parameters of the CNNs for predicting the topic-level labels in Section \ref{sec:image_to_topic} with the modified cost function, to model the image-to-text representation instead of classifying images into categories.
The newly modified output layer has $512$ nodes for bi-grams as $256$ nodes for each word in a bi-gram.





\subsection{Key-Word Generation from Images and Discussion}
\label{sec:key-word-generation}

For any key image in testing, first we predict its topics of three levels (document-level, document-level sub-topics, sentence-level) using the three deep CNN models of \cite{simonyan2014very} in Section \ref{sec:image_to_topic}.
Top 50 key-words in each LDA document-topics are mapped into the word-to-vector space of multivariate variables in $\mathbb{R}^{256\times1}$ (Section \ref{subsec:word2vec}).
Then, the image is mapped to a $\mathbb{R}^{256\times2}$ output vector using the bi-gram CNN model in Section \ref{sec:bi-gram-cnn}.
Lastly, we match each of the $50$ topic key-word vectors of $\mathbb{R}^{256\times1}$ against the first and second half of the $\mathbb{R}^{256\times2}$ output vector using cosine similarity. 
%
The closest key-words at three levels of topics (with the highest cosine similarity against either of the bi-gram words) are kept per image.

\begin{figure}[t]
\begin{center}
   \includegraphics[width=.7\linewidth]{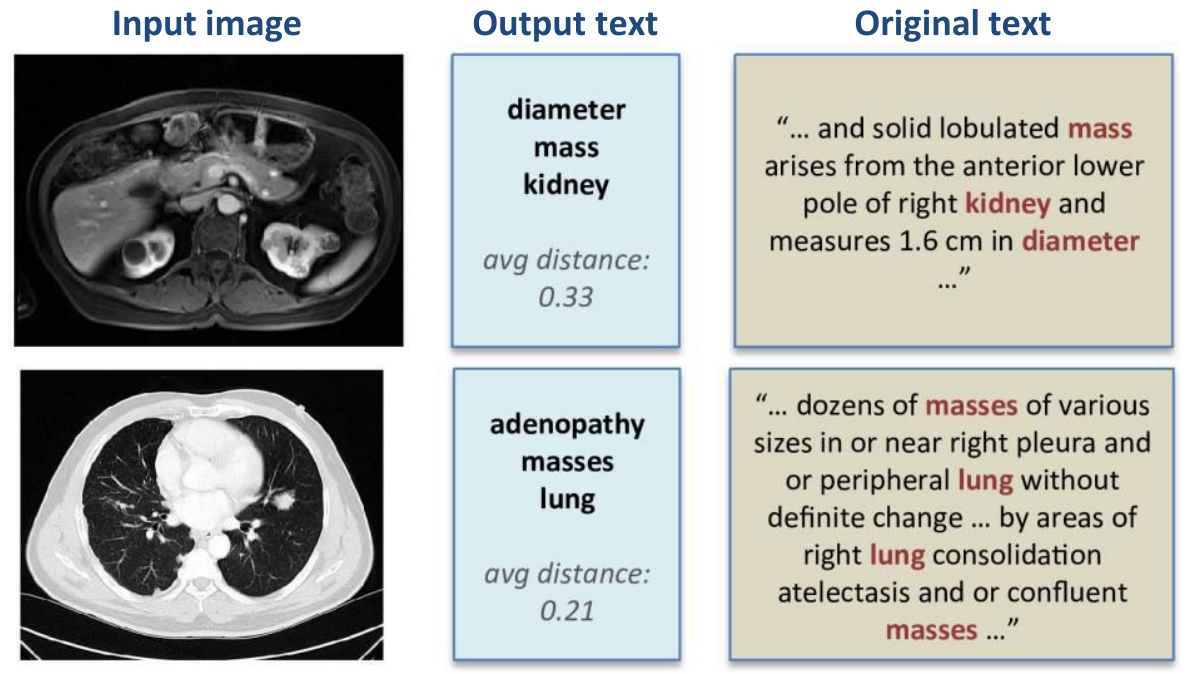} 
\end{center}
   \caption{Examples of text key-word generation results, and average cosine distances between the generated words from the disease-related words in the original texts. The word ``diameter'' appears in the original radiology report of the first image, but not much can be derived by the word only. The rate of predicted disease-related words matching the actual words in the report sentences (recall-at-K, K=1 (R$@$1 score)) was 0.56.}
\label{fig:examples_good}
\end{figure}

The rate of predicted disease-related words matching the actual words in the report sentences (recall-at-K, K=1 (R$@$1 score)) was 0.56.
Two examples of key-word generation are shown in Figure \ref{fig:examples_good}, with three key-words from three categorization levels per image.
We only report R$@$1 score on disease-related words compared to the previous works of \cite{karpathy2014deep,Frome2013devise}, where they report from R$@$1 up to R$@$20 on the entire image caption words (e.g. R$@$1=0.16 on Flickr30K dataset by \cite{karpathy2014deep}).
As we used NLP to parse and extract image-describing sentences from the whole radiology reports, our ground-truth image-to-text associations are much noisier than the caption dataset used by \cite{Frome2013devise,karpathy2014deep}.
Also for that reason, our generated image-to-text associations are not as exact as the generated descriptions by \cite{Frome2013devise,karpathy2014deep}.

\subsubsection{Discussion}

Generating key-words for images by CNN regression shows good feasibility for automated interpretation of patient images.
The generated key-words describe what to expect from the given image, although sometimes unrelated words can be generated too.
Finding and understanding the relations between the generated words will be the next step to explore, for example via more thorough text mining using sophisticated NLP parsing as by \cite{Li2011Composing} and combining them with the specific frequent disease prediction in the next section.

\section{Predicting Presence or Absence of Frequent Disease Types}
\label{sec:disease_prediction}

While the key-words generation in Section \ref{sec:image_to_text} can aid the interpretation of a patient scan, the generated key-words, e.g. ``spine'', ``lung'', are not very specific to a disease in an image.
Nonetheless, one of the ultimate goal for large-scale radiology image/text analysis would be to automatically diagnose disease from a patient scan.
In order to achieve the goal of automated disease detection, we added an additional pipeline of mining disease words rather than disease-related words using radiology semantics, and predicting these in an image using CNNs with softmax cost-function.

\subsection{Mining Presence/Absence of Frequent Disease Terms}

The disease names in Disease Ontology (DO) contains not only disease terms but also non-disease terms as well describing a disease.
Some examples of disease names in DO containing non-disease terms are ``occlusion of gallbladder'' (DOID: 9714), ``acute diarrhea'' (DOID: 0050140), ``strawberry gallbladder'' (DOID: 10254), and ``exocrine pancreatic insufficiency'' (DOID: 13316).
Nonetheless, it is rare that ``occlusion of gallbladder'' or ``exocrine pancreatic insufficiency'' is described in radiology reports exactly that way, making it difficult to mine specific disease terms with presence or absence.

The Unified Medical Language System (UMLS) of \cite{lindberg1993unified,humphreys1998unified} integrates and distributes key terminology, classification and coding standards, and associated resources to promote creation of more effective and inter-operable biomedical information systems and services, including electronic health records.
It is a compendium of many controlled vocabularies in the biomedical sciences, created in 1986 and maintained by the National Library of Medicine.

The Metathesaurus (\cite{schuyler1993umls}) forms the base of the UMLS and comprises over 1 million biomedical concepts and 5 million concept names, where all of them are collected from the over 100 incorporated controlled vocabularies and classification systems.
The Metathesaurus is organized by concept, where each concept has specific attributes defining its meaning and is linked to the corresponding concept names. 
The Metathesaurus has 133 semantic types that provide a consistent categorization of all concepts represented in it.
Among the 133 semantic types we chose to focus on ``T033: finding'' and ``T047: disease or syndrome'', as they seemed most relevant to be disease specific.
Examples of some other semantic types we did not focus on this study are: ``T017: anatomical structure'', ``T074: medical device'', and ``T184: sign or symptom''.

RadLex (\cite{langlotz2006radlex}) is a unified language to organize and retrieve radiology imaging reports and medical records.
While the Metathesaurus has a vast resource of biomedical concepts, we also used RadLex to confine our disease-term-mining more specifically to radiology related terms.
The mined words are one word terms appearing in the ``T033: finding'' and ``T047: disease or syndrome'' of the UMLS Metathesaurus appearing also in RadLex (RadLex is not a subset of Metathesaurus).

\begin{figure}[t]
\begin{center}
   \includegraphics[width=1.0\linewidth]{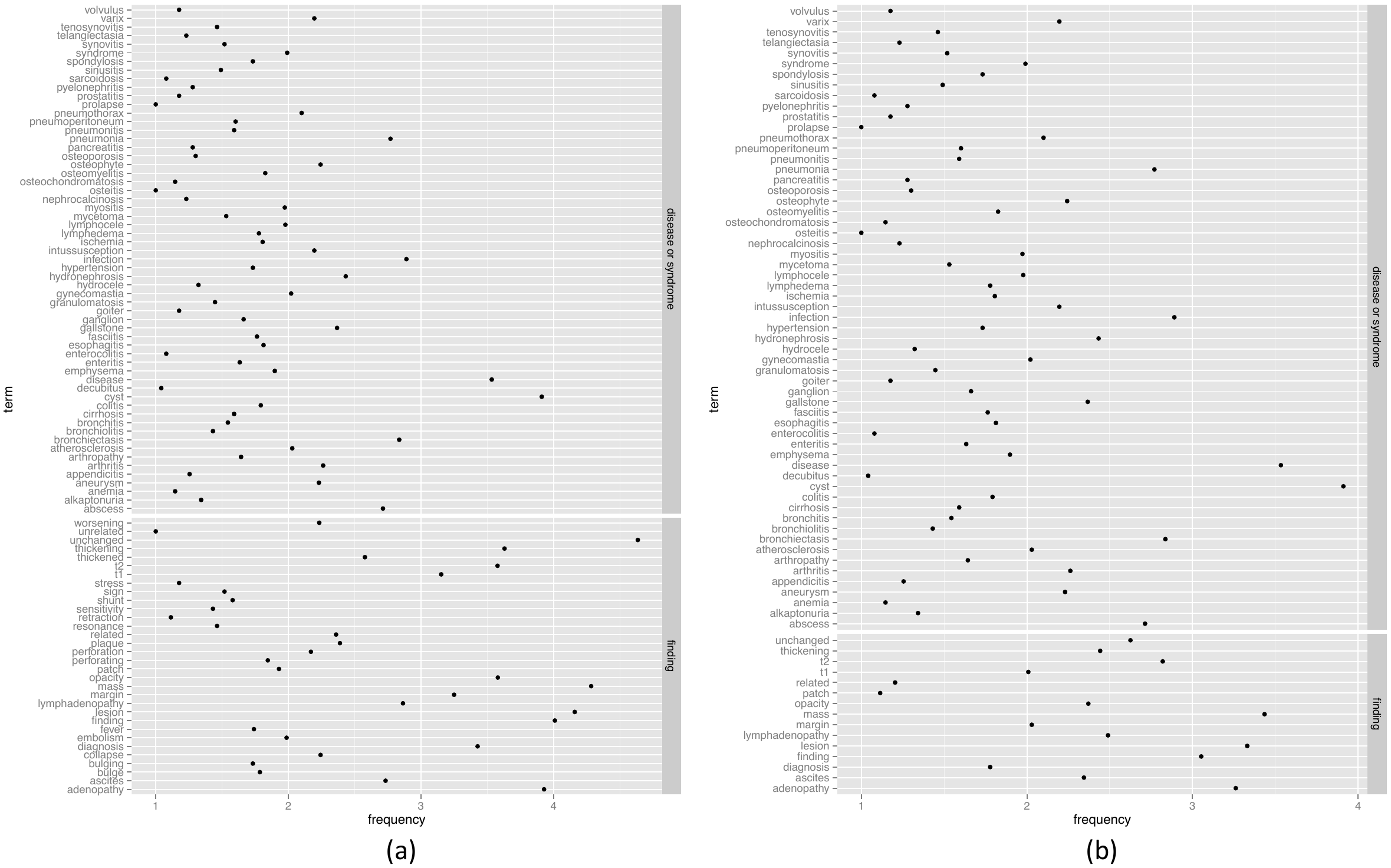}
\end{center}
   \caption{Number of occurrences (frequencies) of semantic terms ``T033: finding'' and ``T047: disease or syndrome'' in UMLS Metathesaurus and also appearing in RadLex, detected as (a) assertion and (b) negation in the radiology reports. Frequencies are shown in $\log_{10}$ scale.}
\label{fig:asss_negs_t033_t047}
\end{figure}

We are not only interested in disease terms associated with image, but also whether the disease mentioned is present or absent.
After detecting semantic terms of ``T033: finding'' and ``T047: disease or syndrome'', we used the assertion/negation detection algorithm of \cite{chapman2001simple,chapman2013extending} to detect presence and absence of disease terms.
The algorithm of \cite{chapman2001simple,chapman2013extending} locates trigger terms which can indicate a clinical condition as negated or possible and determines which text falls within the scope of the trigger terms.
The number of occurrences ``T033: finding'' and ``T047: disease or syndrome'' detected as assertion or negations in radiology reports are shown in Figure \ref{fig:asss_negs_t033_t047}.

While the assertion/negation detection of ``T047: disease or syndrome'' seemed specific enough, the detection of ``T033: finding'' was not.
For example, it seemed difficult to derive any specific disease information from 43,219 occurrences of possible ``unchanged'' and 422 occurrences of negated ``unchanged''.
Some other similar examples are: 10,236 occurrences of possible ``finding'' and 1,129 occurrences of negated ``finding''; 3,781 occurrences of possible ``t2'' (a MRI image modality) and 661 occurrences of negated ``t2''.
We therefore decided to focus on ``T047: disease or syndrome'' terms only, and further ignored the terms which occurred less then 10 times in the whole radiology reports.
The total number of ``T047: disease or syndrome'' terms for detecting their presence are 59, and the total number of the terms for detecting their absence are 18.

\subsection{Predicting Disease in Images using CNN}

Similarly to the object detection task in the ImageNet challenge, we match and detect disease terms found in the sentence of radiology reports referring the image using CNN and softmax cost function.
Softmax models the probability of an instance being the class $j$ of $N$ total classes with normalized exponential.
It is a generalization of the logistic function summarizing an $N$ dimensional vector $\mathbf{z}$ of real values in the range of 0 to 1:

\begin{equation}
  \sigma(\mathbf{z})_j = \frac{e^{z_j}}{\sum_{k=1}^{N} e^{z_k}}.
\label{eq:softmax}
\end{equation}

\noindent
Softmax is often implemented at the final layer of the neural networks used for classification, and is a standard among the CNN based approaches in ImageNet object recognition challenge.

In addition to assigning disease terms to images, we also assign negated disease terms as absence of the diseases in the images.
The total number of labels is 77 (59 present, 18 absent).
If more than one disease term is mentioned for a image, we simply assigned the terms multiple times for an image.
Some statistics on the number of assertion/negation occurrences per image are shown in Table \ref{tab:image_to_asss_negs_stats}.

\begin{table}
\begin{center}
\resizebox{1.0\linewidth}{!}{
\begin{tabular}{|c|c||c|c||c|c||c|c|}
\hline
 \multicolumn{2}{|c||}{\# images} & \multicolumn{2}{c||}{per image mean/std} & \multicolumn{2}{c||}{\# assertions per image} & \multicolumn{2}{c|}{\# negations per image} \\
\hline\hline
total matching     & 18291  & \# assertions mean & 1.05 & 1/image  & 16133 & 1/image & 1581 \\
\hline
total not matching & 197495 & \# negations mean  & 1.05 & 2/image & 613 & 2/image & 84 \\
\hline
with assertions    & 16827  & \# assertions std  & 0.23 & 3/image   & 81 & 3/image & 0 \\
\hline
with negations     & 1665   & \# negations std   & 0.22 & 4/image  & 0   & 4/image & 0 \\
\hline
\end{tabular}
}
\end{center}
\caption{Some statistics of images-to-disease presence/absence label matching.}
\label{tab:image_to_asss_negs_stats}
\end{table}

As we found in Section \ref{sec:transfer_learning} that transfer learning from the most related model is helpful, we fine-tune the image-to-topic CNN model for the disease prediction model.
For this task we fine-tuned from the image to sentence-level-topic (h3) model in Section \ref{sec:image_to_topic}, as the image-to-sentence-level-topic seems to be most closely related to the image-to-disease-specific-terms model.
Similarly to Section \ref{sec:image_to_topic}, 85\% of image-label pairs were used for training, 5\% for cross-validation, and 10\% for testing.

\subsection{Prediction Result and Discussion}

\begin{figure}
\begin{center}
   \includegraphics[width=1.0\linewidth]{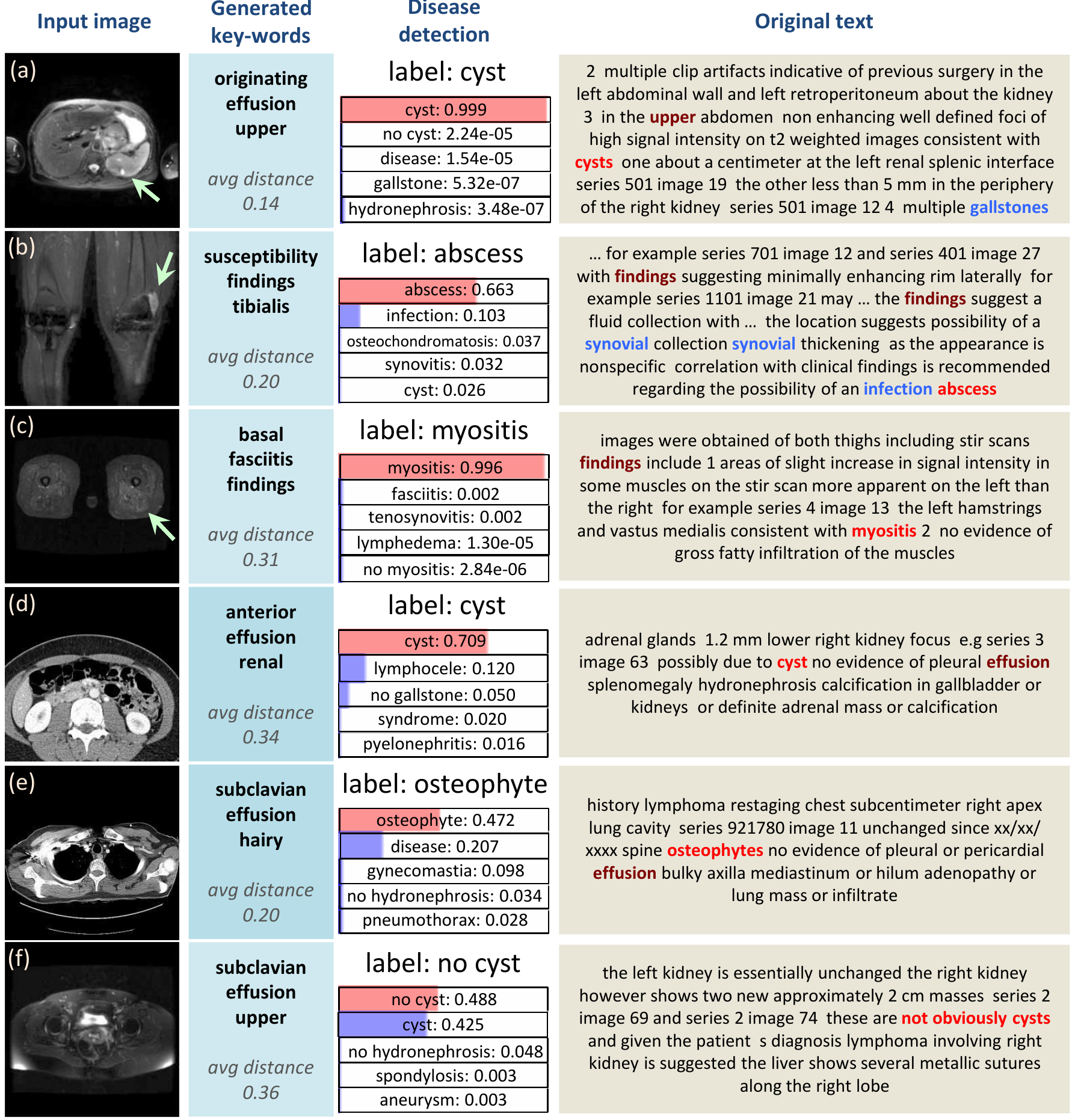}
\end{center}
   \caption{Some examples of final outputs for automated image interpretation, where top-1 probability matches the originally assigned label. Generated key-words appearing in the original text in radiology reports mentioning the image are shown in bold brown, specific disease words assigned as label mentioned in the reports are shown in bold red, and disease words predicted with top-5 probability in the reports are shown in bold blue. The probability assigned to the originally assigned label is shown with a red bar, and the other top-5 probabilities are shown with blue bars. Disease region identified in an image is pointed by arrow.}
\label{fig:final_output_positives}
\end{figure}

With the CNN trained to model image to disease presence/absence prediction, the top-1 test accuracy achieved is 0.71, and top-5 accuracy is 0.88.
We combine this with the previous image-to-topic mapping and key-word generation (Section \ref{sec:key-word-generation}) to generate the final output for comprehensive image interpretation.
Some examples of test cases where top-1 probability output matches the originally assigned disease labels are shown in Figure \ref{fig:final_output_positives}.
It is noticeable that specific disease words are detected with high probability when there is one disease word per image, and with somewhat low top-1 probability for one disease word and the other words within the top-5 probabilities (Figure \ref{fig:final_output_positives} (b) -- `` ... infection abcess'').

We can also notice that automatic label assignment to images can sometimes be challenging.
In Figure \ref{fig:final_output_positives} (d) ``cyst'' is assigned as the correct label based on the original statement ``... possibly due to cyst ...'', but it would be unclear whether cyst will be present on the image (and cyst is not visibly apparent).
It applies similarly to Figure \ref{fig:final_output_positives} (e) where the presence of ``ostephyte'' is not clear from the referring sentence but was assigned as the correct label (and osteophyte is not visibly apparent on the image).
In Figure \ref{fig:final_output_positives} (f) ``no cyst'' was labeled and predicted correctly, but it would be less clear what to derive from the prediction result indicating an absence of a disease then a presence.

\begin{figure}
\begin{center}
   \includegraphics[width=1.0\linewidth]{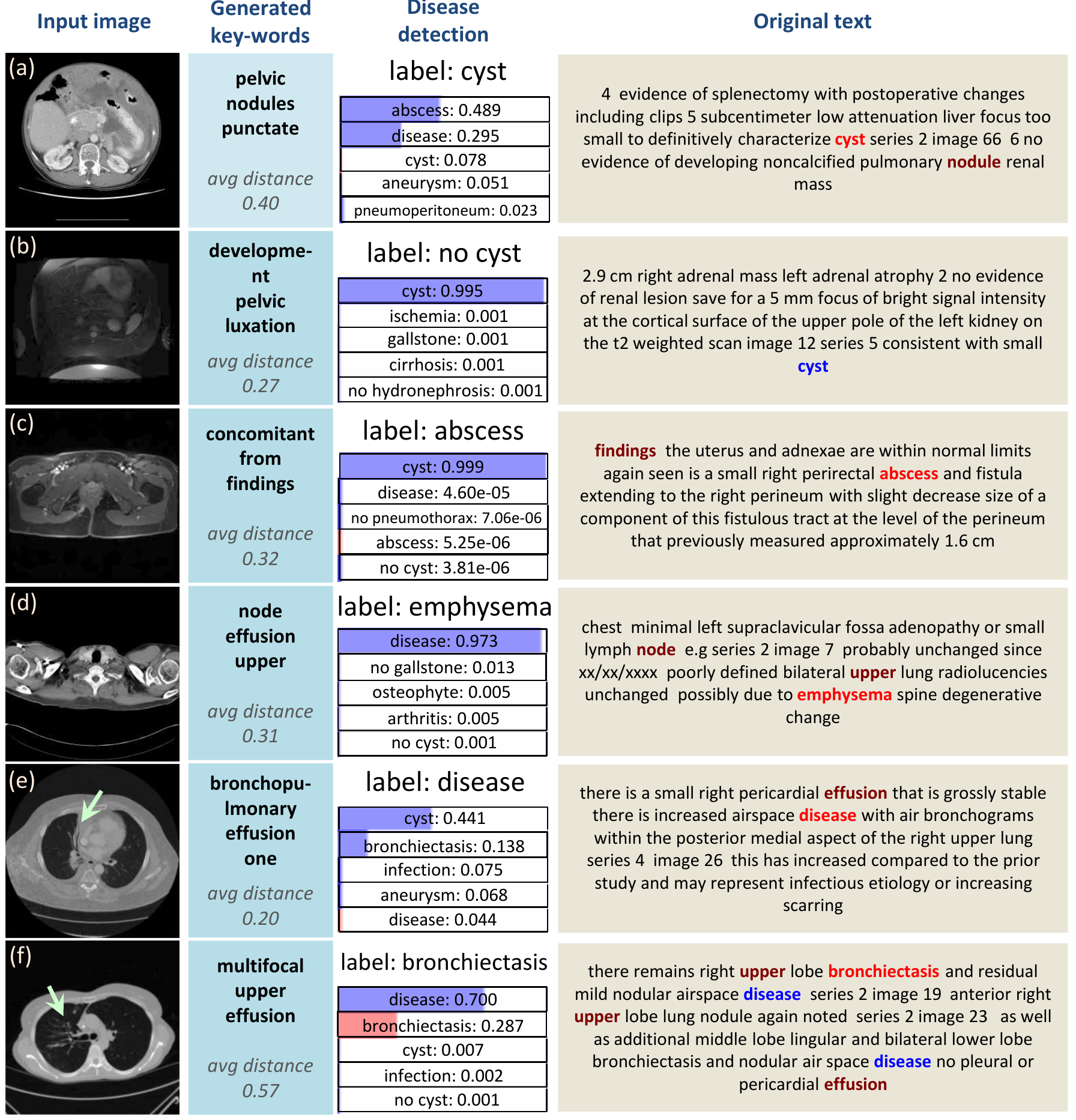}
\end{center}
   \caption{Some examples of final outputs for automated image interpretation where top-1 probability does not match the originally assigned label. One of the top-5 probabilities match the originally assigned labels in the examples of images (a), (c), (d), and (f). None of the top-5 probabilities match the originally assigned labels in the examples of image (b) and (d). However, label assignment of second row example is incorrect, as a failed case of assertion/negation detection algorithm used. Nonetheless, the CNN predicted ``true'' label correctly (``cyst'').}
\label{fig:final_output_negatives}
\end{figure}

Some examples of test cases where top-1 probability does not match the originally assigned labels are shown in Figure \ref{fig:final_output_negatives}.
Four ((a),(c),(e),(f)) of the six examples contain the originally assigned label in the top-5 probability predictions, which is coherent with the relatively high (88\%) top-5 prediction accuracy.

Here again, Figure \ref{fig:final_output_negatives} (a) was automatically labeled as ``cyst'', but cyst is not clearly visible on the image where the original statement ``... too small to definitely characterize cyst ...'' supports this.
The example of Figure \ref{fig:final_output_negatives} (b) shows a failed case of assertion/negation algorithm, where ``cyst'' is detected as negated based on the statement ``... small cyst''.
Nonetheless, true label (``cyst'') is detected as its top-1 probability.
For Figure \ref{fig:final_output_negatives} (c) ``cyst'' is predicted where the true label assigned was ``abscess'', however cyst and abscess are sometime visible similar.
Similarly to Figure \ref{fig:final_output_positives} (d) it is unclear to find emphysema from the statement like `` ... possibly due to emphysema'' (and emphysema is not visibly present), but it would be challenging to correctly interpretate such statement for label assignment.
Figure \ref{fig:final_output_negatives} (e) shows a disease which can be bronchiectasis, however it is also unclear from the image.
Nonetheless, bronchiectasis is predicted with the second highest probability.
Bronchiectasis is visible on Figure \ref{fig:final_output_negatives} (f), and it was predicted with second highest probability too.

\subsubsection{Discussion}

Automated mining of disease specific terms enables us to predict disease more specifically with promising result.
However, compared to image-to-topic modeling in Section \ref{sec:image_to_topic} where image labeling was based on topic modeling and loose coupling of image-to-keyword pairs, by matching the images to more specific disease words we lose about 90\% of the images for the analysis due to nonspecific original statements.
The proportion of the cases where radiologists indicate a disease as strongly positive or negative is often much less then the cases where they describe a finding rather vaguely.
By mining and assigning the semantic label ``T033: finding'' will yield us more image to specific-disease-label pairs.
However, it is probably less specific to model an image with a generic term as ``mass'' (which is a more vague indication of a specific disease such as ``cyst'' or ``tumor'') and detecting it than modeling and detecting an image with a more specific term as ``cyst'' (similarly to ``finding'' or ``unchanged'').

It is a compromise between whether to go for big data and loose labels, or to go for smaller data and more accurate labels.
The key-word generation from the rather loose labeling scheme enables us to use most of the available 216K images.
While the generated key-words can help understand the contents of the image, sometimes they are not specific and can also be irrelevant.
More specific mining and assignment of specific disease labels to image could provide more accurate and precise disease prediction, however only about 10\% of the total images are matched by this scheme.
Another alternative is to obtain annotation by radiologists to be even more specific, but the amount of data available will be even smaller due to the time and cost limitations.

Utilizing bigger data will enable us to make a more generalizable model, but labeling will become more challenging as the amount of data gets bigger and becomes more heterogeneous.
The compromise between the amount of data and the quality of labels seems to be a recurring dilemma probably in the majority of the automated mining in big data applications.
More advanced NLP processing and comprehensive analysis of hospital discharge summaries, progress notes, and patient histories might address the need to get more specific information relating to an image even when the original image descriptions are not very specific.

\section{Conclusion}

It has been unclear how to extend the significant success in image classification using deep convolutional neural networks from computer vision to medical imaging.
What are the clinically relevant image labels to be defined, how to annotate the huge amount of medical images required by deep learning models, and to what extent and scale the deep CNN architecture is generalizable in medical image analysis are the open questions.

In this paper, we present an interleaved text/image deep mining system to extract the semantic interactions of radiology reports and diagnostic key images at a very large, unprecedented scale in the medical domain.
Images are classified into different levels of topics according to their associated documents, and a neural language model is learned to assign disease terms to predict what is in the image.
In addition, we mined and matched specific disease terms for more specific automated image labeling, and demonstrated promising results.

To the best of our knowledge, this is the first study performing a large-scale image/text analysis on a hospital picture archiving and communication system database.
Our report-extracted key image database is the largest one ever reported and is highly representative of the huge collection of radiology diagnostic semantics over the last decade.
Exploring effective deep learning models on this database opens new ways to parse and understand large-scale radiology image informatics. 

We hope that this study will inspire and encourage other institutions in mining other large unannotated clinical databases, to achieve the goal of establishing a central training resource and performance benchmark for large-scale medical image research, similar to the ImageNet of \cite{deng2009imagenet} for computer vision.


\acks{This work was supported in part by the Intramural Research Program of the National Institutes of Health Clinical Center, and in part by a grant from the KRIBB Research Initiative Program (Korean Visiting Scientist Training Award), Korea Research Institute of Bioscience and Biotechnology, Republic of Korea.
This study utilized the high-performance computational capabilities of the Biowulf Linux cluster at the National Institutes of Health, Bethesda, MD (http://biowulf.nih.gov).
We thank NVIDIA for the K40 GPU donation.}

\vskip 0.2in
\bibliography{shin_cvpr2015}

\end{document}